\documentclass{article}
\usepackage[T1]{fontenc}
\usepackage{iclr2027_conference,times}
\usepackage{amsmath,amssymb,amsthm,mathtools}
\usepackage{graphicx,booktabs,tabularx,array,flafter}
\newcolumntype{Y}{>{\raggedright\arraybackslash}X}
\usepackage{algorithm,algpseudocode}
\usepackage{tikz}
\usetikzlibrary{arrows.meta,positioning,fit,backgrounds}
\usepackage[hidelinks]{hyperref}
\usepackage{url}
\usepackage{enumitem}
\newcommand{\method}{\textsc{DAAF}}
\newcommand{\E}{\mathbb{E}}

\newcommand{\C}{\mathcal{C}}
\newcommand{\D}{\mathcal{D}}
\newcommand{\ind}{\mathbf{1}}

\DeclareMathOperator{\BCE}{BCE}
\newtheorem{proposition}{Proposition}

\title{DAAF: From Failure Localization to Editable System Assets in LLM Agents}
\iclrfinalcopy
\author{%
\hfill\begin{tabular}[t]{c}
Xiaoyang Yuan\thanks{Equal contribution.}, Qi Liu\footnotemark[1], Yubin Ruan, Xinyi Mou, Zhuomeng Zhang, Wenjin Wang, \\
Hanying Jiao, Di Wu, Mingye Xu, Yi Bin, Ke Feng, Zixun Sun \\[2pt]
Tencent
\end{tabular}\hfill\mbox{}%
}
\begin{document}
\maketitle
% arXiv preprint: remove the ICLR conference header line and rule.
\lhead{}\chead{}\rhead{}
\fancyhead{}
\renewcommand{\headrulewidth}{0pt}
\begin{abstract}
Deployed LLM agents increasingly rely on persistent, versioned system assets such as routing rules, knowledge segments, prompt instructions, and reusable skills. Failure-localization methods can identify where an error manifests in an agent or execution trace, but repair requires a different decision: \emph{which editable system asset should be changed, and is that change expected to improve the task outcome?} We study this gap through component-attribute failure attribution, where diagnosis targets versioned, addressable items rather than execution locations. We propose the \textbf{Detection-Aware Attribution Framework (DAAF)}, which learns the effects of valid attribute replacements and amortizes this intervention evidence into deployment-time diagnosis. DAAF combines sparse and noisy failure signals to decide whether intervention is warranted, learns component-type-conditioned replacement effects from controlled replays evaluated by executable task outcomes, and shares supervision across requests with compatible intervention responses. At diagnosis time, DAAF uses only the observed execution, registered candidates, and available failure signals; it requires neither counterfactual replay nor task reward and returns \texttt{no\_change}, a repair target, or an unresolved decision when evidence is insufficient. On held-out $\tau^2$-bench Telecom tasks, DAAF achieves 80.72\% attribute Hit@1, recovers 62.65\% of failed executions while limiting clean-task regression to 3.23\%, and reaches 71.93\% overall task success. These results show that intervention-grounded attribute attribution can connect failure localization to executable system repair.
\end{abstract}

\section{Introduction}
\label{sec:intro}

LLM agents increasingly operate as persistent systems rather than isolated single-turn programs. A deployed agent may serve many requests through shared routing rules, retrieval knowledge, prompt instructions, runtime policies, and reusable skills~\citep{yao2023react,wu2024autogen,wang2024voyager}. As these assets evolve with data, policies, and dependencies, failures may arise from system state that persists across requests rather than from a single input. Existing failure-localization methods identify responsible agents, execution steps, or functional modules and provide useful evidence for debugging and recovery~\citep{zhang2025whowhen,zhu2025agentdebug,zhu2026agentdebugx}. Multi-agent studies likewise identify system design and coordination as recurring sources of failure~\citep{cemri2025mast}. However, localization alone does not determine which persistent system element should be changed. Its output is typically an agent identity, step index, or module label~\citep{zhang2025whowhen,chen2026elephant,zhang2025agentracer,zhu2025agentdebug}, which need not correspond to a versioned, addressable object that can be modified and independently re-evaluated. A retrieval step, for example, may execute correctly while using stale knowledge selected by an upstream rule. Effective repair therefore requires identifying not only where a failure appears, but which concrete system change is likely to improve the outcome.

Consider an information-retrieval agent that repeatedly returns outdated policy information. Possible repairs include replacing the knowledge segment, changing the selection rule, or revising the instruction. These changes act on different persistent objects, may affect other requests that reuse them, and can have beneficial, neutral, harmful, or interaction-dependent effects. Selecting among them therefore requires evidence about what happens when the objects are actually modified, consistent with recent intervention- and counterfactual-based approaches to agent debugging~\citep{ma2025dover,shah2026car,bonagiri2026causalflow}. We study whether such intervention outcomes, collected during training, can predict for a new execution whether a change is needed and which registered edit is most likely to help.

This problem raises three practical challenges. First, failure evidence is incomplete: trajectory anomalies, user feedback, and disagreement across repeated executions may all be informative, but their availability and reliability vary across requests~\citep{kuhn2023semantic,farquhar2024entropy,pathak2025silent,mehta2026consistency}. Second, edits are heterogeneous and interacting. Changing a knowledge segment and changing a runtime setting follow different constraints, yet their effects must be compared using the same task-level outcome. Third, repair evidence must transfer selectively across requests: two executions may touch the same assets without sharing the same failure cause, so evidence should be shared only when their contexts and intervention responses are compatible.

We propose the \textbf{Detection-Aware Attribution Framework (DAAF)} to address these challenges. DAAF treats the expected task-level effect of changing a registered editable asset as the attribution target. Multi-Signal Weakly-Supervised Detection (MSWSD) combines available failure signals while representing unavailable observations explicitly. Component-Type Conditional Attribution (CTCA) learns signed effects of registered replacements from controlled replays evaluated by executable task outcomes, using type-conditioned predictors for different edit roles. To capture interactions, CTCA estimates marginal effects in composed replacement settings rather than only from isolated edits. DAAF further builds fixed training neighborhoods from requests with compatible contexts and intervention responses to share repair evidence across related executions. At diagnosis time, MSWSD determines whether intervention is warranted and CTCA scores the registered candidates, returning \texttt{no\_change}, a single registered repair, or an unresolved outcome. No task evaluator or counterfactual replay is required during diagnosis, and DAAF does not generate replacement content.

On held-out $\tau^2$-bench Telecom tasks~\citep{barres2025tau2bench}, DAAF achieves 80.72\% attribute Hit@1, recovers 62.65\% of failed executions while limiting clean-task regression to 3.23\%, and reaches 71.93\% overall task success. Controlled intervention analysis further shows that non-zero replacement effects increase from 11.36\% under isolated edits to 31.02\% in composed settings, while training-derived conditional effects remain informative for unseen repair decisions.

Our main contributions are:
\begin{itemize}[leftmargin=1.5em,itemsep=1pt,topsep=2pt,parsep=0pt]

    \item We introduce \emph{component-attribute failure attribution}, which targets versioned, editable system assets rather than execution locations and defines attribution through the signed task-level effect of valid replacements.

    \item We propose \textbf{DAAF}, which combines weak-signal failure detection, intervention-supervised component-type-conditioned effect prediction, interaction-aware replay targets, and cross-request consistency. Controlled interventions provide offline supervision, while diagnosis requires only the observed execution, available failure signals, and registered candidates.

    \item We demonstrate that attribute-level diagnosis translates into executable repair. DAAF reaches 80.72\% attribute Hit@1 and 62.65\% failure recovery with 3.23\% clean-task regression on held-out Telecom tasks, while controlled analyses support interaction-aware attribution and transfer of intervention evidence to unseen repairs.

\end{itemize}
\section{Related Work}
\label{sec:related}

\noindent\textbf{Failure detection from weak signals.}
Semantic uncertainty and semantic entropy use variation across generated answers as evidence of model uncertainty~\citep{kuhn2023semantic,farquhar2024entropy}. In agent executions, trajectory anomalies and behavioral inconsistency provide additional signals of failure~\citep{pathak2025silent,mehta2026consistency}. Data programming and Snorkel combine multiple noisy labeling sources~\citep{ratner2016dataprogramming,ratner2017snorkel}, while FlyingSquid studies efficient estimation from latent weak sources~\citep{fu2020flyingsquid}. DAAF uses this line of work for failure detection, but detection is only the first stage of the repair problem. MSWSD combines signals with different availability and reliability to decide whether an execution should enter the repair stage; the subsequent attribution problem is supervised by executed interventions rather than weak failure evidence alone.

\noindent\textbf{Failure attribution and intervention-based debugging.}
Existing failure-attribution methods primarily localize responsible agents, execution steps, or other execution-level causes~\citep{zhang2025whowhen,chen2026elephant,zhang2025agentracer,wang2026chief}. Who\&When predicts responsible agents and decisive steps~\citep{zhang2025whowhen}; TraceElephant and CHIEF incorporate broader execution context and hierarchical causal structure~\citep{chen2026elephant,wang2026chief}. AgenTracer constructs attribution supervision through counterfactual replay and fault injection~\citep{zhang2025agentracer}, OAT scores anomalous steps using dynamics learned from successful trajectories~\citep{yeh2026oat}, and AgentDebug and AgentDebugX connect structured failure diagnosis with recovery~\citep{zhu2025agentdebug,zhu2026agentdebugx}. CAR, CausalFlow, and DoVer use counterfactual replay or targeted interventions to identify, validate, or repair failure-inducing execution decisions~\citep{shah2026car,bonagiri2026causalflow,ma2025dover}, while MP-Bench evaluates settings with multiple plausible attributions~\citep{in2026mpbench}. DAAF focuses on a different attribution target: a versioned, addressable system asset whose registered replacement can be executed and evaluated. Controlled interventions provide supervision for the effect of these replacements, and the learned effects are used at diagnosis time without replaying candidate edits for each new request.

\noindent\textbf{Feedback-driven repair and system optimization.}
Reflexion stores feedback in reflective memory~\citep{shinn2023reflexion}, while Self-Refine iteratively revises model outputs using self-feedback~\citep{madaan2023selfrefine}. Work on intrinsic self-correction further examines the limitations of self-generated feedback for improving reasoning~\citep{huang2024selfcorrect}. At the system level, ProTeGi searches over prompt edits using textual feedback~\citep{pryzant2023protegi}, DSPy optimizes parameterized language-model pipelines~\citep{khattab2024dspy}, and TextGrad propagates feedback to system variables~\citep{yuksekgonul2025textgrad}. These system-optimization methods address how prompts or other pipeline variables can be improved once an optimization objective is available. DAAF instead addresses the preceding diagnosis problem: given an observed execution and a registry of editable assets, it estimates which asset has evidence supporting a change. The selected asset can subsequently be passed to an optimizer or human maintainer to construct a new replacement.
\section{DAAF: Learning Which Attributes to Change}
\label{sec:method}

\begin{figure*}[t]
\centering
\includegraphics[width=0.95\textwidth]{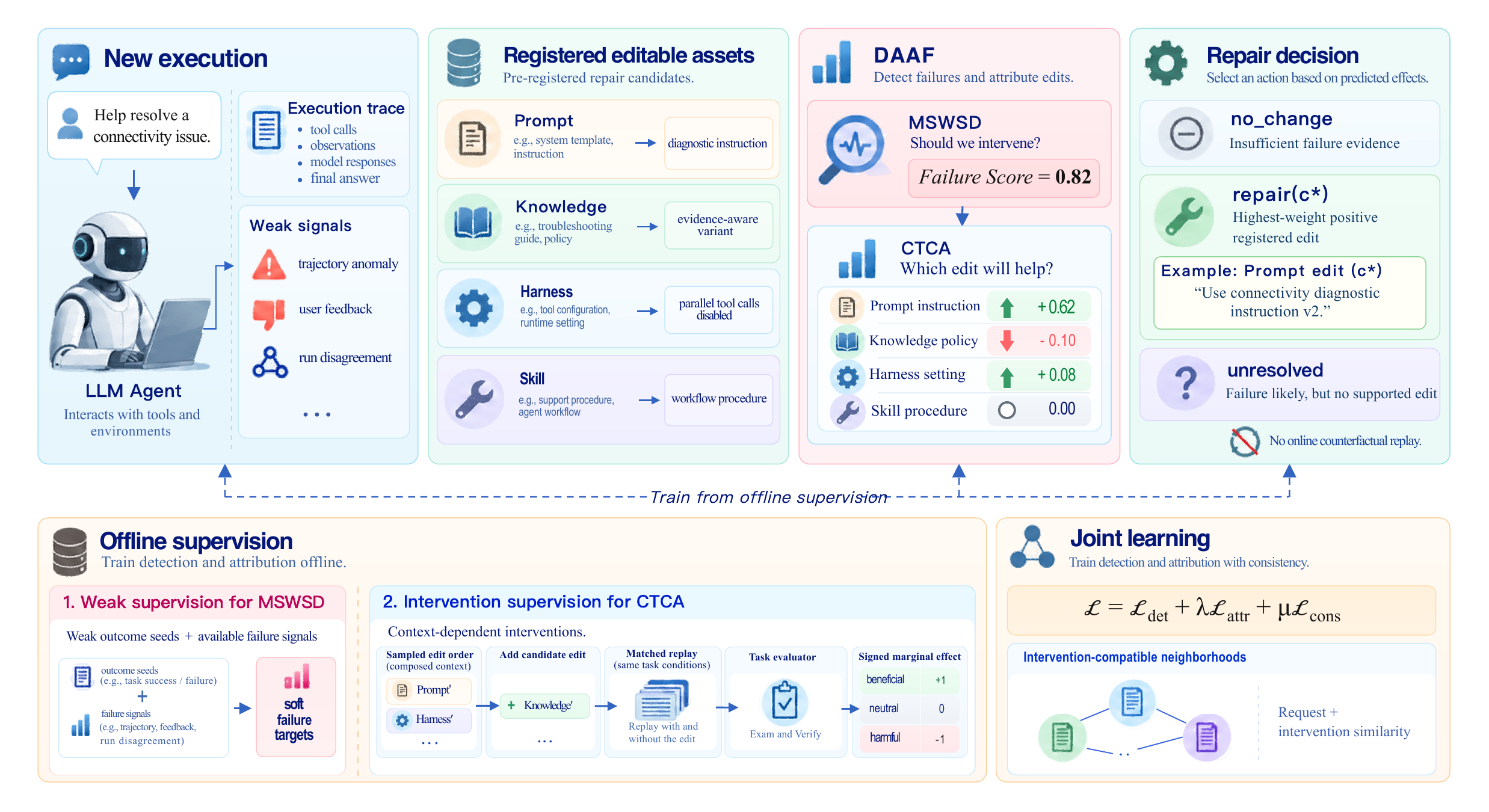}
\caption{
Overview of DAAF.
At diagnosis time (solid arrows), MSWSD estimates whether intervention is warranted and CTCA predicts signed effects for registered edits; their outputs determine \texttt{no\_change}, a single repair, or an unresolved decision.
During training (dashed arrows), weak-label targets supervise failure detection, controlled replacements evaluated by executable task outcomes supervise attribute effects, and intervention-compatible requests provide cross-request consistency.
}
\label{fig:architecture}
\end{figure*}

DAAF formulates agent diagnosis as a repair decision over persistent, editable system assets. Given a new execution, the method first determines whether the observed evidence warrants intervention and then predicts which registered change is expected to improve the task outcome. Training has access to executable environments, so registered replacements can be applied, replayed, and evaluated by task outcomes. Diagnosis does not have this access. It uses only the observed execution, the current system registry, and the failure signals available at that time. The key distinction from execution-level localization is that DAAF predicts the effect of changing an addressable system asset rather than only identifying where a failure appears. Figure~\ref{fig:architecture} summarizes the training and diagnosis pipelines.

DAAF has three learned parts. MSWSD estimates failure evidence from weak deployment signals (Section~\ref{sec:mswsd}). CTCA learns the signed effect of each registered replacement from controlled replays (Section~\ref{sec:ctca}). Joint learning shares repair evidence across compatible requests and combines detection and attribution into an abstaining repair decision (Section~\ref{sec:joint}).

\subsection{Problem Setup and Registered Edits}
\label{sec:method-setup}

Let $A_v$ denote an agent system at version $v$. A request $q$, initial environment state $e$, and execution randomness $\xi$ produce a trajectory
$\tau=A_v(q,e;\xi)$. The diagnostic input is
\begin{equation}
z=(q,\tau,v,\C_v,s,m),
\qquad
s\in[0,1]^J,\quad
m\in\{0,1\}^J,
\label{eq:diagnostic-input}
\end{equation}
where $\C_v$ is the editable registry for system version $v$, $s_i$ is a failure signal, and $m_i$ records whether that signal is available.

\textbf{Editable assets.}
A \emph{component} groups system elements with a common functional role, and an \emph{attribute} is the atomic registered item that DAAF can select for repair. Each attribute $c\in\C_v$ has a stable identifier, component type $t(c)$, source location, current version, and one or more validated replacements. We organize attributes into five roles: Node, Harness, Knowledge, Prompt, and Skill. These correspond to routing or iteration control, execution hooks, retrieval evidence, model instructions, and reusable procedures, respectively. The taxonomy is not required to be exhaustive; DAAF only assumes that candidate attributes are addressable and can be replaced through a validated interface.

\begin{table}[t]
\caption{Examples of registered edits in the Telecom environment. Replacements are fixed before diagnosis and are not generated by DAAF.}
\label{tab:types}
\small
\begin{tabularx}{\linewidth}{@{}lYY@{}}
\toprule
Role & Registered attribute & Admissible replacement \\
\midrule
Skill
& Support procedure
& \texttt{tech\_support\_manual.md} $\rightarrow$ \texttt{tech\_support\_workflow.md} \\

Knowledge
& Policy evidence segment
& Official policy $\rightarrow$ registered policy variant with evidence-handling guidance \\

Prompt
& Instruction / system template
& Official text $\rightarrow$ text with a fixed diagnostic or verification suffix \\

Harness
& Parallel tool-call setting
& Provider default $\rightarrow$ disabled \\
\bottomrule
\end{tabularx}
\end{table}

Table~\ref{tab:types} gives examples from the Telecom environment. All replacements are versioned and validated before training or evaluation. DAAF therefore predicts \emph{which registered asset should be changed}; generating new replacement content is outside the diagnosis procedure.

\textbf{Intervention outcomes.}
For executable training tasks, let $\C(z)\subseteq\C_v$ be the candidate set and let $A_v^{S,\rho}$ denote the system obtained by applying replacement policy $\rho$ to attributes $S\subseteq\C(z)$. Starting from matched task conditions, the outcome of a replacement set is
\begin{equation}
R_z(S)=
\E_{\xi}\!\left[
\operatorname{Eval}\!\left(
A_v^{S,\rho}(q,e;\xi)
\right)
\right].
\label{eq:intervention-reward}
\end{equation}
Here $\operatorname{Eval}$ is an executable task evaluator. In our $\tau^2$-bench experiments~\citep{barres2025tau2bench}, it is the official binary task evaluator rather than an LLM judge. Controlled comparisons of $R_z(S)$ provide supervision about whether a registered change improves, preserves, or harms task success. Invalid or unavailable replays do not produce attribution targets.

\subsection{Weak-Signal Failure Detection}
\label{sec:mswsd}

The first decision is whether an execution should enter the repair stage. DAAF uses a registered set of weak failure signals $\mathcal S$, including trajectory anomalies~\citep{pathak2025silent}, visible user feedback, and disagreement across matched repeated executions when repeated runs are available~\citep{mehta2026consistency}. For signal $i$, $s_i\in[0,1]$ denotes failure evidence, $m_i\in\{0,1\}$ its availability, and $u_i\in[0,1]$ a fixed reliability value. The detector represents unavailable observations explicitly:
\begin{equation}
p_\theta(z)=
\sigma\!\left(
b+
\sum_{i\in\mathcal S}w_i m_i u_i s_i+
\sum_{i\in\mathcal S}v_i(1-m_i)
\right).
\label{eq:detector}
\end{equation}
The output $p_\theta(z)$ estimates the evidence that intervention is warranted. Signal transformations and reliability values are fitted or specified from training data and frozen before evaluation.

Following data programming~\citep{ratner2016dataprogramming,ratner2017snorkel}, noisy seed annotations and labeling functions are combined into soft failure targets $\widetilde y_z$ with confidence $c_z$. Label-model fitting and target generation use separate training folds. The detector is trained with
\begin{equation}
\mathcal L_{\mathrm{det}}=
\frac{1}{|\D_{\mathrm{seed}}|}
\sum_{z\in\D_{\mathrm{seed}}}
c_z\,\BCE(\widetilde y_z,p_\theta(z))
+
\frac{\beta}{|\D_u|}
\sum_{z\in\D_u}
c_z\,\BCE(\widetilde y_z,p_\theta(z)),
\label{eq:detloss}
\end{equation}
where $\D_u$ contains the remaining training interactions and $\beta$ controls their contribution. Development data are used only for calibration and threshold selection.

\subsection{Intervention-Supervised Attribute Effects}
\label{sec:ctca}

Detection indicates that a change may be needed, but it does not identify which change will help. CTCA learns this distinction from the outcomes of controlled interventions.

\textbf{Marginal intervention effects.}
The effect of an edit can depend on other changes already applied to the system. For a uniformly sampled permutation $\pi$ of $\C(z)$, let $P_\pi(c)$ contain the attributes appearing before $c$. We define the attribution target
\begin{equation}
\phi_c^\rho(z)=
\E_{\pi}\!\left[
R_z(P_\pi(c)\cup\{c\})-
R_z(P_\pi(c))
\right].
\label{eq:shapley}
\end{equation}
This Shapley-style marginal effect~\citep{shapley1953value} is estimated with sampled permutations~\citep{castro2009sampling}. Positive values indicate that the registered replacement tends to improve task outcome, while negative values indicate harm. Evaluating edits in composed settings also exposes effects that may be absent when each replacement is tested only against the original system.

\textbf{Predicting effects without replay.}
For each candidate attribute, a shared encoder represents the request, relevant execution context, current attribute content, and replacement description. Component-type-conditioned heads then predict the same signed quantity:
\begin{equation}
a_{\psi,c}(z)\approx\phi_c^\rho(z),
\qquad
a_{\psi,c}(z)\in[-1,1].
\label{eq:effect-prediction}
\end{equation}
For training interactions $\D_I$, let $O_z$ contain the attributes with valid replay-derived targets. CTCA minimizes
\begin{equation}
\mathcal L_{\mathrm{attr}}=
\frac{1}{|\D_I|}
\sum_{z\in\D_I}
\frac{1}{|O_z|}
\sum_{c\in O_z}
\left(
a_{\psi,c}(z)-
\widehat{\phi}_c^{\mathrm{MC}}(z)
\right)^2.
\label{eq:attrloss}
\end{equation}
Unavailable comparisons are omitted rather than treated as zero. Once trained, CTCA predicts replacement effects directly from a new execution and its registered candidates. The task evaluator and intervention replays are used only to construct offline supervision.

\subsection{Joint Learning and Repair Decision}
\label{sec:joint}

The two learned quantities serve different roles. The detector estimates whether intervention is needed, while CTCA estimates how each registered change is expected to affect the task. DAAF also uses replay outcomes to determine when evidence can be shared across requests.

\textbf{Cross-request consistency.}
Executions that touch the same assets do not necessarily benefit from the same repair. We therefore construct training neighborhoods only between cases that share a compatible system version, candidate set, replacement policy, and task evaluator. For an eligible pair,
\begin{equation}
W_{ij}=
\kappa(q_i,q_j)\,\omega_i\omega_j
\exp\!\left(
-\frac{
\|\widehat\phi_i^{\mathrm{MC}}-
\widehat\phi_j^{\mathrm{MC}}\|_2^2
}{2\ell^2}
\right),
\label{eq:neighbors}
\end{equation}
where $\kappa$ measures request compatibility, $\ell$ controls sensitivity to intervention-response differences, and $\omega_i,\omega_j$ are fixed validity weights. The neighborhoods are constructed from training data and frozen before joint optimization.

DAAF converts positive predicted effects into routing weights,
\begin{equation}
r_c=
\frac{[a_{\psi,c}]_+}
{\epsilon_0+\sum_j[a_{\psi,j}]_+},
\qquad
r_\bot=
\frac{\epsilon_0}
{\epsilon_0+\sum_j[a_{\psi,j}]_+},
\qquad
f_c=p_\theta r_c,
\label{eq:routing}
\end{equation}
with unresolved mass $f_\bot=p_\theta r_\bot$. Let $f_{\theta,\psi}(z)$ collect the candidate and unresolved masses. With $Z_W=\sum_{i<j}W_{ij}$, the joint objective is
\begin{equation}
\mathcal L=
\mathcal L_{\mathrm{det}}
+\lambda\mathcal L_{\mathrm{attr}}
+\frac{\mu}{Z_W}
\sum_{i<j}
W_{ij}
\left\|
f_{\theta,\psi}(z_i)-
f_{\theta,\psi}(z_j)
\right\|_2^2,
\label{eq:joint}
\end{equation}
where the consistency term is zero when $Z_W=0$. This term follows the general idea of graph-based regularization~\citep{zhou2003consistency,belkin2006manifold}, but the edges here depend on observed intervention responses in addition to request compatibility. The model therefore shares repair evidence between requests that behave similarly under actual system changes, rather than between requests that are merely similar in text.

\textbf{Diagnosis.}
The primary DAAF policy returns at most one registered edit. Let $\tau_f$ be the development-selected intervention threshold. The final decision is
\begin{equation}
d(z)=
\begin{cases}
\texttt{no\_change},
& p_\theta(z)<\tau_f,\\
\bot,
& p_\theta(z)\ge\tau_f
\ \text{and}
\max_c r_c\le r_\bot,\\
\operatorname{repair}(c^\star),
& \text{otherwise},
\end{cases}
\qquad
c^\star=
\arg\max_{c:\,a_{\psi,c}>0} r_c .
\label{eq:decision}
\end{equation}
Low failure evidence leaves the current system unchanged. When intervention is warranted but no candidate has enough positive support, DAAF returns an unresolved decision. Otherwise it selects the registered edit with the strongest predicted positive effect. Composed repairs are studied separately in Section~\ref{sec:experiments}; they are not part of the primary single-edit diagnosis rule.

The resulting model uses controlled interventions only as offline supervision. Diagnosis requires no task reward, environment replay, or search over candidate replacements. Finite-sample results for repeated signals and replay-based effect estimation are given in Appendix~\ref{app:theory}.
\section{Experiments}
\label{sec:experiments}

We evaluate \method{} from failure diagnosis to executable system repair. $\tau^2$-bench Telecom~\citep{barres2025tau2bench} is our primary benchmark because it provides an executable environment, an official task evaluator, and persistent assets that can be modified and re-executed. We first measure whether attribute diagnosis leads to actual task recovery, then examine the interaction structure used to supervise CTCA and whether training-side intervention effects transfer to unseen tasks. Additional analyses study incomplete failure evidence and cross-request evidence sharing. We use Who\&When~\citep{zhang2025whowhen} as an auxiliary check of conventional trajectory-level localization.

\subsection{Experimental Setup}

\textbf{Benchmarks.}
$\tau^2$-bench Telecom provides an executable customer-service environment with an official task evaluator and persistent system assets. We use it for attribute diagnosis, executed repair, and intervention analysis. Who\&When contains recorded multi-agent trajectories with native agent- and step-level failure labels. Since it does not expose editable system assets, it is used only for trajectory-level localization.

\textbf{Models and task splits.}
Telecom execution, user simulation, and the component-aware LLM diagnostician use GPT-5.6-luna~\citep{openai2026gpt56}. The held-out evaluation set contains all 114 official Telecom default tasks, with 83 natural failures and 31 natural successes under the frozen evaluation configuration. Training, development, and evaluation partitions are separated by canonical base-task identity. Who\&When localization uses DeepSeek-V4-Flash~\citep{deepseek2026v4}; we treat this experiment as an auxiliary localization check rather than a model-matched repair comparison. Full split and model details are given in Appendix~\ref{app:settings}.

\textbf{Training supervision.}
MSWSD is trained from 476 trajectories over 395 tasks using weak outcome seeds and diagnosis-time signals. CTCA uses 220 training tasks with five registered attributes. Two sampled replacement orders and two matched repetitions produce 5{,}280 controlled executions and 1{,}100 signed marginal-effect targets. A disjoint 50-task development set is used only for model selection and calibration.

\textbf{Evaluation protocol.}
Attribute Hit@1 measures whether the top-ranked edit belongs to the independently validated repair set. Joint F1 also accounts for unnecessary interventions. Failure Recovery is the fraction of natural failures that succeed after repair, Clean Regression is the fraction of natural successes that become failures, and Overall Success measures final success over all 114 tasks. Repair results use three matched execution repeats with shared task--repeat seeds across methods and report mean $\pm$ standard deviation. Metric definitions, repair-label validation, paired confidence intervals, and matched-seed construction are detailed in Appendix~\ref{app:metrics} and Appendix~\ref{app:matched-seeds}.

\subsection{Attribute Diagnosis and Executed Repair}
\label{sec:repair}

We begin with the main question: does DAAF identify an editable asset whose modification improves the underlying task? No-edit Rerun measures recovery from rerunning the unchanged system. Component-aware LLM receives the full execution trace and descriptions of the registered candidates and directly selects an edit. LLM + MSWSD adds the learned failure detector while retaining the same LLM-based edit proposal. Full \method{} combines failure detection with intervention-supervised prediction of replacement effects.

\begin{table}[htbp]
\centering
\caption{
End-to-end results on the 114 held-out $\tau^2$-bench Telecom tasks.
Attribute Hit@1 is evaluated on 83 natural failures; Joint F1 also includes the 31 natural successes.
Repair metrics are mean $\pm$ standard deviation over three matched execution repeats.
}
\label{tab:telecom-main}
\small
\setlength{\tabcolsep}{4.2pt}
\begin{tabularx}{\linewidth}{@{}Yccccc@{}}
\toprule
Method
& \shortstack{Attribute\\Hit@1 $\uparrow$}
& \shortstack{Joint\\F1 $\uparrow$}
& \shortstack{Failure\\Recovery $\uparrow$}
& \shortstack{Clean\\Regression $\downarrow$}
& \shortstack{Overall\\Success $\uparrow$} \\
\midrule

No-edit Rerun
& -- & --
& $12.05 \pm 3.62$
& $26.88 \pm 1.86$
& $28.66 \pm 2.20$ \\

Component-aware LLM
& 40.96
& 51.13
& $35.74 \pm 4.56$
& $19.35 \pm 3.23$
& $47.95 \pm 2.53$ \\

LLM + MSWSD
& 36.14
& 47.62
& $33.73 \pm 4.82$
& $24.73 \pm 3.72$
& $45.03 \pm 2.68$ \\

\method{}
& \textbf{80.72}
& \textbf{68.02}
& $\mathbf{62.65 \pm 3.19}$
& $\mathbf{3.23 \pm 3.23}$
& $\mathbf{71.93 \pm 1.52}$ \\
\bottomrule
\end{tabularx}
\end{table}

\textbf{Attribute diagnosis translates into task recovery.}
\method{} reaches 80.72\% Attribute Hit@1, compared with 40.96\% for the component-aware LLM. Uniform selection over the validated candidate sets gives 37.35\% expected Hit@1, so the improvement cannot be explained by the presence of multiple acceptable repairs alone (Appendix~\ref{app:extended-localization}). The difference remains visible after the selected edits are executed. \method{} repairs $62.65\pm3.19\%$ of the original failures, compared with $35.74\pm4.56\%$ for the component-aware LLM and $12.05\pm3.62\%$ for rerunning the unchanged system. Overall Success reaches $71.93\pm1.52\%$. The main result therefore connects attribute-level diagnosis to realized task recovery rather than evaluating attribution only against diagnostic labels.

\textbf{Failure recognition is not sufficient for repair selection.}
Adding MSWSD to the LLM proposal process does not improve repair over the component-aware LLM. A stricter detector-only control keeps the DAAF repair pipeline fixed but removes replacement-effect supervision; its Failure Recovery falls to $4.42\pm2.51\%$, 58.23 percentage points below Full \method{}, and Clean Regression rises to $24.73\pm1.86\%$ (Appendix~\ref{app:detection-only}). Conversely, attaching CTCA to the original LLM proposal interface reaches only 34.94\% Hit@1 and 46.03\% Joint F1 (Appendix~\ref{app:extended-localization}). These controls indicate that intervention-derived effect supervision is important within the joint diagnosis and attribution decision, rather than failure detection or replacement scoring alone being sufficient.

\textbf{The learned effects improve repair selectivity without online candidate search.}
\method{} reduces Clean Regression to $3.23\pm3.23\%$, compared with $19.35\pm3.23\%$ for the component-aware LLM. The higher recovery is therefore not obtained by applying more edits indiscriminately. The learned effect estimates distinguish beneficial replacements from neutral or harmful alternatives. The 5{,}280 intervention executions used to train CTCA are collected offline; diagnosis does not replay all five candidates for each new request. Under the evaluation protocol, Full \method{} uses 2.00 online environment executions per natural failure, with the detailed accounting reported in Appendix~\ref{app:online-cost}.

\subsection{Interaction-Aware Attribution and Transfer}
\label{sec:interaction}

CTCA is trained on marginal effects from composed replacement settings because the effect of one edit can depend on other changes already present in the system. We first test whether this dependence is substantial in the replay corpus, then examine whether conditional effects estimated only from training interventions remain informative on unseen Telecom tasks.

\textbf{Replacement effects depend strongly on intervention context.}
Across all five registered attributes, 11.36\% of standalone replacements change the official task outcome, compared with 31.02\% in composed configurations. Four attributes that are nearly inactive in isolation, with activity rates of 0.53--1.25\%, affect 21.34--30.85\% of composed states. Standalone probing finds only 59 of the 515 task--attribute pairs that exhibit a measurable effect somewhere in their replacement chain, missing 88.54\% of the observed effective pairs. For 472 of the 1{,}100 task--attribute targets, sampled marginal contributions differ in sign across replacement orders. These results support modeling replacement effects as context-dependent rather than assigning each edit a fixed effect from isolated behavior. Per-attribute statistics are reported in Appendix~\ref{app:interaction-stats}.

We next rank candidate second edits using conditional effects computed exclusively from the training replay corpus. Held-out outcomes are not used to determine this ordering.

\begin{table}[htbp]
\centering
\caption{
Transfer of training-derived conditional replacement effects.
Each candidate is applied after the Full DAAF repair on held-out Telecom tasks.
$\Delta$ Recovery is measured in percentage points relative to Full DAAF.
}
\label{tab:interaction-transfer}
\small
\setlength{\tabcolsep}{4.0pt}
\begin{tabularx}{\linewidth}{@{}Ycccc@{}}
\toprule
Second edit
& \shortstack{Train Conditional\\Effect (pp)}
& \shortstack{Failure\\Recovery $\uparrow$}
& \shortstack{$\Delta$ Recovery\\95\% CI}
& \shortstack{Clean\\Regression $\downarrow$} \\
\midrule

Prompt: instruction
& $+6.79$
& $\mathbf{81.53 \pm 2.51}$
& $\mathbf{+18.88\ [10.84,26.91]}$
& $\mathbf{2.15 \pm 3.72}$ \\

Prompt: system template
& $+1.71$
& $64.26 \pm 0.70$
& $+1.61\ [-4.02,7.23]$
& $6.45 \pm 3.23$ \\

None (Full \method{})
& $0.00$
& $62.65 \pm 3.19$
& reference
& $3.23 \pm 3.23$ \\

Harness: parallel tool calls
& $-1.72$
& $62.25 \pm 2.78$
& $-0.40\ [-6.83,6.02]$
& $4.30 \pm 1.86$ \\

Knowledge: main policy
& $-5.02$
& $60.64 \pm 1.84$
& $-2.01\ [-8.03,6.02]$
& $8.60 \pm 1.86$ \\
\bottomrule
\end{tabularx}
\end{table}

\textbf{Training-side conditional effects remain informative on held-out tasks.}
The held-out recovery values follow the same ordering as the training-derived conditional effects in this five-policy comparison. Prompt Instruction has the largest positive training effect and raises Failure Recovery from $62.65\%$ to $81.53\%$. The gain is 18.88 percentage points with a 95\% confidence interval of $[10.84,26.91]$, while Clean Regression decreases to $2.15\pm3.72\%$. The other candidates follow the same ordering, although their recovery differences are smaller and their confidence intervals include zero. These results provide evidence that intervention structure estimated from the training replay corpus can transfer to unseen repair decisions without using held-out outcomes to construct the ranking.

\subsection{Supporting Analyses}
\label{sec:components}

\textbf{Unavailable evidence should not be treated as observed negative evidence.}
Zero-filling unavailable signals reduces AUPRC from 75.77\% to 71.31\% and AUROC from 56.06\% to 44.58\% on the natural cohort. On the balanced development set, the corresponding scores fall from 66.00\%/73.92\% to 50.00\%/48.00\%. Mean imputation matches the missingness-aware representation on both reported splits. The supported conclusion is therefore that missing evidence should be distinguished from an observed zero, rather than that explicit masks dominate all imputation strategies. Full comparisons are reported in Appendix~\ref{app:incomplete-evidence}.

\textbf{Intervention responses provide additional structure for cross-request sharing.}
Intervention-response neighborhoods reach 62.63\% held-out agreement, compared with 60.09\% for request-similarity neighborhoods and 58.59\% for random edges. The intervention-based construction has the highest agreement on all ten attribute hold-out splits (Appendix~\ref{app:neighborhood-quality}). The improvement is modest in absolute terms but consistent across splits, suggesting that observed responses to system changes provide information beyond lexical request similarity when selecting cases from which repair evidence should be shared.

\subsection{Conventional Failure Localization}
\label{sec:localization}

Who\&When does not expose editable system assets, so this experiment is not used to support the executable repair claims above. Instead, it checks whether the trajectory frontend retains conventional agent- and step-level localization capability. Table~\ref{tab:whowhen} reports the benchmark's native Step and Joint metrics; reporting details are given in Appendix~\ref{app:whowhen}.

\begin{table}[htbp]
\centering
\caption{
Native Who\&When~\citep{zhang2025whowhen} accuracy (\%, higher is better).
Step measures exact step-level agreement; Joint requires both agent and step.
}
\label{tab:whowhen}
\small
\begin{tabularx}{\linewidth}{@{}Ycccc@{}}
\toprule
& \multicolumn{2}{c}{Algorithm}
& \multicolumn{2}{c}{Handcrafted} \\
\cmidrule(lr){2-3}\cmidrule(l){4-5}
Method
& Step & Joint
& Step & Joint \\
\midrule

All-at-Once
& 22.22 & 12.70
& 1.72 & 1.72 \\

Step-by-Step
& 30.16 & 30.16
& 24.14 & 22.41 \\

Binary Search
& 0.79 & 0.79
& 12.07 & 12.07 \\

AgentDebugX~\citep{zhu2026agentdebugx}
& 42.06 & 42.06
& 18.97 & 17.24 \\

\method{} frontend
& \textbf{48.02}
& \textbf{48.02}
& \textbf{25.86}
& \textbf{24.14} \\
\bottomrule
\end{tabularx}
\end{table}

\textbf{The DAAF frontend retains trajectory-level localization capability.}
Under this reporting setup, the DAAF frontend reaches 48.02\% Step and Joint accuracy on Algorithm traces, compared with 42.06\% for AgentDebugX. On Handcrafted traces, it reaches 25.86\% Step and 24.14\% Joint accuracy, compared with 18.97\% and 17.24\%, respectively. We treat these results as an auxiliary compatibility check because Who\&When does not provide editable-asset interventions and the comparison is not a model-matched executable repair evaluation. The repair and intervention claims of DAAF are established by the Telecom experiments.
\section{Conclusion}

DAAF addresses the gap between locating an agent failure and identifying which persistent system asset should be changed. It connects weak failure evidence to versioned, editable repair targets by learning the effects of registered replacements from controlled interventions, using these outcomes as offline supervision while requiring no per-candidate replay or task reward at diagnosis time. On held-out $\tau^2$-bench Telecom tasks, DAAF achieves 80.72\% attribute Hit@1, recovers 62.65\% of failed executions, and limits clean-task regression to 3.23\%. Controlled replay analysis further shows that non-zero replacement effects increase from 11.36\% under isolated edits to 31.02\% in composed settings, supporting context-dependent attribution rather than fixed isolated edit relevance. Together, these results show that intervention-grounded attribution can connect failure localization to executable repair decisions over persistent system assets. The current formulation assumes a registered candidate set, leaving automatic asset discovery and replacement generation for future work.
\label{maintextend}
\section*{Reproducibility Statement}
\label{sec:reproducibility-statement}

We provide the information needed to reproduce the DAAF training and evaluation pipeline throughout the paper and appendices. Appendix~\ref{app:signals} specifies failure-signal extraction, weak supervision, and intervention-target construction. Appendix~\ref{app:registry} describes the editable-asset registry, registered replacements, replay semantics, attribution representation, and replay budget. Appendix~\ref{app:learning} reports the learning procedure and final configuration, including Table~\ref{tab:full-config} and Algorithm~\ref{alg:train}. Appendix~\ref{app:theory} gives the estimation results and proofs, while Appendix~\ref{app:metrics} defines the evaluation metrics, independent repair-label validation, matched execution seeds, paired confidence intervals, and conditional-effect analysis. Additional controls and implementation settings are provided in Appendices~\ref{app:additional-results} and~\ref{app:settings}. We will release the training and evaluation code, model checkpoints, registered-asset configurations, and Telecom replay artifacts where redistribution permits.

\section*{Ethics Statement}

This work studies failure diagnosis and system repair in controlled benchmark environments and does not involve live deployment or human-subject experiments. Controlled interventions modify registered assets only within executable evaluation environments and do not affect production systems or real users. Existing benchmark data are used under their original access and licensing conditions, and released artifacts will follow applicable redistribution requirements. Because changes to persistent agent assets may affect many subsequent requests, results in these controlled environments should not be interpreted as sufficient evidence for unsupervised deployment in safety-critical settings. Practical deployment of automated repair systems should include appropriate validation, access control, monitoring, and human oversight.

\section*{AI Use Statement}

Generative AI tools were used as research assistants in this work. ChatGPT~\citep{openai2026chatgpt} and other generative AI systems assisted with literature-metadata checks, manuscript drafting and language editing, mathematical derivation checks, code debugging and optimization, and experimental analysis. Generative models are also used within the experimental pipeline: LLM-assisted weak annotations contribute to the failure-detection supervision described in Appendix~\ref{app:signals}, and language models serve as agents, simulators, or diagnosticians under the settings reported in Appendix~\ref{app:settings}. The core research problem, methodological contributions, intervention design, and experimental protocol were developed by the authors. All AI-assisted text, code, analyses, and generated annotations were reviewed by the authors, who take responsibility for the final content, claims, references, and reported results.

\bibliography{references}
\bibliographystyle{iclr2027_conference}

\appendix
\appendix

\section{Signal Extraction and Supervision}
\label{app:signals}

\textbf{Behavioral disagreement.}
For $K\ge2$ repeated runs of the same request and starting state, let
$d(\tau_i,\tau_j)\in[0,1]$ compare normalized final answers and action sequences. We define
\[
s_1=\binom{K}{2}^{-1}\sum_{i<j}d(\tau_i,\tau_j).
\]
Answer normalization follows the motivation of semantic uncertainty: semantically equivalent outcomes should receive compatible representations
\citep{kuhn2023semantic,farquhar2024entropy}. Semantic comparison and action alignment are fixed on training data. If only one matched run is available, disagreement is treated as unavailable rather than assigned a default value.

\textbf{Trajectory anomaly.}
The structural anomaly signal is
\[
s_2=0.5r_{\mathrm{err}}+0.3r_{\mathrm{rep}}+0.2e_{\mathrm{empty}},
\]
where $r_{\mathrm{err}}$ is the fraction of visible tool results containing an explicit error,
$r_{\mathrm{rep}}$ is the fraction of repeated tool calls with identical names and arguments beyond their first occurrence, and
$e_{\mathrm{empty}}$ indicates an empty final response.
The anomaly reliability is fixed to $1.0$. Evidence records retain the source events and observed recovery, so an intermediate anomaly does not by itself imply final task failure.

\textbf{User feedback.}
The feedback signal uses visible outcome reports, corrections, negative reactions, and abandonment events. The extractor first uses the latest goal-specific outcome report. If none is recognized, it falls back to the latest recognized correction, negative reaction, or abandonment event. Contradictory outcome evidence yields no feedback signal. Table~\ref{tab:feedback-map} gives the fixed score and reliability mapping used in the Telecom experiments.

\begin{table}[htbp]
\centering
\caption{Frozen user-feedback mapping used by MSWSD. Larger scores indicate stronger failure evidence.}
\label{tab:feedback-map}
\small
\begin{tabularx}{0.78\linewidth}{@{}Ycc@{}}
\toprule
Feedback event & Failure score & Reliability \\
\midrule
Explicit positive outcome & 0.00 & 0.75 \\
Explicit negative outcome & 1.00 & 0.75 \\
Correction & 0.75 & 0.50 \\
Explicit negative reaction & 0.60 & 0.40 \\
Abandonment & 0.50 & 0.30 \\
\bottomrule
\end{tabularx}
\end{table}

\textbf{Weak outcome supervision.}
Each training seed records its source and confidence. Telecom outcome seeds are LLM-assisted weak annotations anchored to available task-outcome evidence. A confidence-weighted logistic label model with pairwise signal-interaction features is fitted by canonical base-task folds and produces cross-fitted soft targets $\widetilde y_z$ with confidence $c_z$. These targets are frozen before fitting the detector. The detector represents each available signal through its reliability-weighted value and an explicit availability indicator, so an unavailable signal is not treated as an observed zero.

\textbf{Intervention supervision for CTCA.}
CTCA uses the official executable task outcome defined in Eq.~\ref{eq:intervention-reward}. For each sampled replacement order, adjacent coalitions are replayed from matched task conditions and their outcome difference provides one signed marginal sample for the added attribute. The Monte Carlo average of these samples estimates the target in Eq.~\ref{eq:shapley}. Positive, zero, and negative effects are retained. Official task outcomes are used only to construct offline attribution targets and are never exposed as diagnostic features at evaluation time.

\section{Registry, Interventions, and Attribution Heads}
\label{app:registry}

\textbf{Registry representation.}
Each editable attribute is stored with an identifier, component type, system version, content hash, source locator, dependency identifiers, allowed values, and replacement-policy identifier. Structural locators are preferred to bare line numbers so that an attributed item maps back to a persistent system asset. Inseparable changes are represented as one intervention unit, while dependencies between separable attributes remain explicit.

\textbf{Telecom registered assets.}
Table~\ref{tab:telecom-assets} summarizes the five attributes used in the controlled Telecom study. All assets are pinned to the corresponding upstream version and file hashes in the registry.

\begin{table}[htbp]
\centering
\caption{Registered Telecom attributes and replacement roles used for intervention supervision.}
\label{tab:telecom-assets}
\small
\begin{tabularx}{\linewidth}{@{}lYY@{}}
\toprule
Type & Registered attribute & Replacement role \\
\midrule
Prompt
& Instruction
& Adds explicit diagnostic ordering, distinguishes agent-side from user-side actions, and requests user verification before advancing. \\

Prompt
& System template
& Adds end-of-task verification and requires unresolved issues to be stated explicitly rather than inferred from tool completion. \\

Harness
& Parallel tool calls
& Changes the runtime setting from provider-default behavior to disabled parallel tool calls. \\

Knowledge
& Main policy
& Adds an evidence-handling section to the registered Telecom policy variant. \\

Skill
& Tech-support procedure
& Replaces the manual-style support description with the registered workflow-style procedure. \\
\bottomrule
\end{tabularx}
\end{table}

\textbf{Replacement semantics.}
Harness replacements change compatible execution rules; Knowledge and Prompt replacements operate on versioned semantic segments with matched roles; Skill replacements preserve the required inputs, outputs, and dependency ordering. Candidate sets are validated before replay so that every sampled coalition has a well-defined executable interpretation. The registry restricts each intervention to its predeclared original and replacement values.

\textbf{Replay semantics.}
Marginal comparisons begin from matched starting conditions and share execution seeds when the environment permits. Telecom interventions replay each task from its initial state because a registered attribute can affect the first agent decision. Every valid replay records the coalition, replacement identities, seed, and official outcome. A fixed retry policy handles transient execution failures before a target is marked unavailable.

\textbf{Attribution representation.}
CTCA uses fixed SHA-256 token hashing after NFKC normalization and lowercasing. Tokens are mapped to signed 64-dimensional buckets and L2-normalized. For each candidate attribute, the model concatenates a 64-dimensional request block, a 64-dimensional visible-execution block, a 64-dimensional attribute/replacement block, eight numeric features, and a five-way component-type indicator, producing a 205-dimensional input row. The visible-execution representation contains the planner context, executed decisions, tool observations, final reply, and bounded visible diagnostic evidence. The attribute block contains current content, replacement content, interface information, and dependencies.

\textbf{Attribution heads.}
A 32-unit shared hidden representation feeds five type-specific signed-effect heads. The encoder is initialized with variance scaled by the input dimension; typed heads use small random initialization with a fixed bias. The signed output is bounded to $[-1,1]$. The Telecom attribution model contains 6,757 trainable parameters.

\textbf{Replay budget.}
A complete replacement permutation visits $d+1$ nested coalitions and supplies one marginal sample per attribute. With $M$ sampled replacement orders and $R$ matched repetitions per visited coalition, the uncached budget is $M(d+1)R$. The Telecom CTCA corpus uses $d=5$, $M=2$, and $R=2$, yielding 24 valid replay executions per training task and 5,280 valid executions over 220 tasks. These executions provide offline supervision; diagnosis does not enumerate the candidate replacements online.

\section{Learning and Optimization Details}
\label{app:learning}

\textbf{Detection supervision.}
For seeded interactions $\D_{\mathrm{seed}}$ and remaining training interactions $\D_u$, the confidence-weighted detector loss is
\begin{equation}
\mathcal L_{\mathrm{det}}=
\frac{1}{|\D_{\mathrm{seed}}|}\sum_{z\in\D_{\mathrm{seed}}}c_z\BCE(\widetilde y_z,p_\theta(z))
+\frac{\beta}{|\D_u|}\sum_{z\in\D_u}c_z\BCE(\widetilde y_z,p_\theta(z)).
\label{eq:detloss-app}
\end{equation}
A term is omitted when its set is empty. Cross-fitted targets and reliability values remain fixed during subsequent joint training.

\textbf{Attribution supervision.}
For cases $\D_I$ with at least one observed replay target, let $O_z$ denote the observed attributes. The implementation uses a fixed case weight $\omega_z^{\mathrm{attr}}$:
\begin{equation}
\mathcal L_{\mathrm{attr}}=
\frac{1}{|\D_I|}\sum_{z\in\D_I}\frac{\omega_z^{\mathrm{attr}}}{|O_z|}
\sum_{c\in O_z}\big(a_{\psi,c}(z)-\widehat\phi_c^{\mathrm{MC}}(z)\big)^2.
\label{eq:attrloss-app}
\end{equation}
Observed zero and negative effects remain in the loss. In the final Telecom configuration, the case weight follows the target-weight $\times$ weak-label-confidence rule reported in Table~\ref{tab:full-config}.

\textbf{Routing and abstention.}
The implementation follows Eq.~\ref{eq:routing}. The final inference setting uses $\epsilon_0=0.01$.

\textbf{Intervention-compatible neighborhoods.}
Eligible training pairs share a compatible system version, candidate set, replacement policy, and task evaluator. For such a pair,
\begin{equation}
W_{ij}=\kappa(q_i,q_j)\,\omega_i\omega_j
\exp\!\left(-\frac{\|\widehat\phi_i^{\mathrm{MC}}-\widehat\phi_j^{\mathrm{MC}}\|_2^2}{2\ell^2}\right),
\label{eq:neighbors-app}
\end{equation}
where $\kappa$ is the nonnegative cosine similarity of fixed request features and $\omega_i$ summarizes supervision reliability. The graph retains the five highest-weight neighbors per case, is symmetrized, and is frozen before joint optimization. We use $\ell=0.2$.

\textbf{Final configuration.}
Table~\ref{tab:full-config} reports the development-selected configuration used for the Telecom results. The development partition is used only for model selection and calibration.

\begin{table}[htbp]
\centering
\caption{Final DAAF configuration used for the Telecom experiments.}
\label{tab:full-config}
\small
\begin{tabularx}{0.82\linewidth}{@{}Yl@{}}
\toprule
Setting & Value \\
\midrule
Training seed & 20260924 \\
Base-task folds & 3 \\
Candidate budget & 12 \\
Feature / hidden dimension & 64 / 32 \\
Training epochs / warm-up & 200 / 100 \\
Learning rate & 0.2 \\
Detector weight $\beta$ & 1.0 \\
Attribution weight $\lambda$ & 1.0 \\
Consistency weight $\mu$ & 0.1 \\
Failure threshold & 0.7449 (dev-selected) \\
Temperature & 2.0 \\
Routing $\epsilon_0$ & 0.01 \\
Neighborhood scale $\ell$ & 0.2 \\
Neighbors per case & 5 \\
Gradient clipping & global norm 5.0 \\
Attribution heads & type-specific \\
Attribution weighting & target weight $\times$ weak-label confidence \\
Supervision scorer & official task reward \\
\bottomrule
\end{tabularx}
\end{table}

\begin{algorithm}[t]
\caption{DAAF training and diagnosis}
\label{alg:train}
\begin{algorithmic}[1]
\Require Training traces, noisy seeds, registry, replacement policy $\rho$, replay budget $B$
\State Group splits by canonical base-task identity
\State Fit signal transforms and generate cross-fitted weak targets
\State Warm-start the detector; fix the task evaluator and replacement policy
\State Sample replacement orders and matched replay executions within $B$
\State Estimate signed targets $\widehat\phi^{\mathrm{MC}}$ and train the typed attribution heads
\State Build fixed neighborhoods from request context and intervention responses
\For{each joint training round}
  \State Update $\theta$ using $\mathcal L_{\mathrm{det}}+\mu\mathcal L_{\mathrm{cons}}$, holding $\psi$ fixed
  \State Update $\psi$ using $\lambda\mathcal L_{\mathrm{attr}}+\mu\mathcal L_{\mathrm{cons}}$, holding $\theta$ fixed
\EndFor
\State Apply development-set selection and calibration
\State Diagnose new traces with frozen models using Eq.~\ref{eq:decision}; return \texttt{no\_change}, $\operatorname{repair}(c^\star)$, or $\bot$
\end{algorithmic}
\end{algorithm}

\section{Estimation Properties and Optimization Details}
\label{app:theory}

Two finite-sample effects arise in DAAF. The detector may aggregate repeated observations of noisy failure signals, while CTCA estimates attribute effects from a finite number of sampled intervention orders. The following results characterize how these estimates concentrate as their sampling budgets increase. They provide qualitative finite-sample guidance and are not used to select the experimental budgets.

\begin{proposition}[Repeated-signal concentration]
\label{prop:signal}
For a fixed request and availability mask $m$, let
$S^{(1)},\ldots,S^{(n)}\in[0,1]^J$
be independent signal vectors with common mean $\nu$. Coordinates within each vector may be dependent. Define
\[
\overline S=\frac{1}{n}\sum_{r=1}^{n}S^{(r)},
\qquad
\alpha_i=w_i m_i u_i,
\qquad
b_m=b+\sum_{i=1}^{J}v_i(1-m_i),
\]
with $\alpha$ and $b_m$ fixed independently of
$S^{(1)},\ldots,S^{(n)}$.
Let
\[
p_n=\sigma(b_m+\alpha^\top\overline S),
\qquad
p_\infty=\sigma(b_m+\alpha^\top\nu).
\]
For $\|\alpha\|_1>0$ and $t>0$,
\[
\Pr\!\left(|p_n-p_\infty|\ge t\right)
\le
2\exp\!\left(-\frac{32nt^2}{\|\alpha\|_1^2}\right).
\]
\end{proposition}

The result characterizes sampling variation around the detector output obtained from the mean signal vector. If $p^*$ denotes the true failure probability and $\gamma=|p_\infty-p^*|$ is a separate calibration error, then deviations beyond $\gamma+t$ satisfy the same tail bound.

\begin{proposition}[Finite-budget attribution estimation]
\label{prop:attr}
For a fixed request with $d$ registered candidate attributes, suppose each attribute has $M$ independent unbiased marginal-contribution samples in $[-1,1]$. Let $\widehat\phi^{\mathrm{MC}}$ denote the corresponding sample mean and $\phi^\rho$ the expected marginal effect under replacement policy $\rho$. Then, for any $0<\delta<1$, with probability at least $1-\delta$,
\[
\|\widehat\phi^{\mathrm{MC}}-\phi^\rho\|_\infty
\le
\sqrt{\frac{2\log(2d/\delta)}{M}}.
\]
\end{proposition}

The bound decreases as $O(M^{-1/2})$ and depends only logarithmically on the number of candidate attributes. At small replay budgets the worst-case bound can be loose, so it should not be interpreted as a guarantee that closely spaced effects are distinguishable. This finite-sample uncertainty is consistent with retaining an unresolved outcome when no candidate receives sufficient positive support. In the replay-budget notation of Appendix~\ref{app:registry}, $M$ indexes sampled replacement orders and $R$ indexes matched execution repetitions; the proposition abstracts the resulting marginal estimates into independent bounded samples.

\subsection{Proof of Proposition~\ref{prop:signal}}
Let $X_r=\alpha^\top S^{(r)}$. Its range has width at most $B=\|\alpha\|_1$, even when the $J$ coordinates of a vector are dependent. Independence is required only across vectors. Hoeffding's bounded-sample inequality~\citep{hoeffding1963} gives
\begin{equation}
\Pr\left(\left|\frac1n\sum_r X_r-\E X_r\right|\ge u\right)
\le 2\exp(-2nu^2/B^2).
\end{equation}
Since $\sup_x|\sigma'(x)|=1/4$, $|p_n-p_\infty|\le|\alpha^\top(\overline S-\nu)|/4$. Substituting $u=4t$ proves the bound. If $B=0$, $p_n=p_\infty$ exactly. The triangle inequality yields $|p_n-p^*|\le|p_n-p_\infty|+\gamma$ when a separate calibration bound $\gamma$ is available. Learned coefficients are conditioned on a disjoint training sample.\hfill$\square$

\subsection{Proof of Proposition~\ref{prop:attr}}
For attribute $c$, let $X_{c,r}$ be an unbiased sampled marginal contribution. Since the executable outcome is bounded in $[0,1]$, each paired difference used in Eq.~\ref{eq:shapley} lies in $[-1,1]$. For each fixed $c$,
\begin{equation}
\Pr\left(\left|M^{-1}\sum_{r=1}^M X_{c,r}-\phi_c^\rho\right|\ge t\right)
\le2\exp(-Mt^2/2).
\end{equation}
A union bound over $d$ attributes and setting $2d\exp(-Mt^2/2)=\delta$ proves the result. Independence across different attributes is unnecessary.\hfill$\square$

\textbf{From target estimation to learned heads.}
For any fixed case,
\begin{equation}
\|a_\psi-\phi^\rho\|_\infty\le
\|a_\psi-\widehat\phi^{\mathrm{MC}}\|_\infty+
\|\widehat\phi^{\mathrm{MC}}-\phi^\rho\|_\infty.
\end{equation}
The first term is predictive approximation error and the second is finite-budget replay-estimation error. If the two largest true effects are separated by more than twice the total uniform error, their ranking is preserved.

\subsection{A Descent-Compatible Alternating Variant}
\label{app:optimization}
Freeze labels, replay targets, reliability weights, candidate sets, and neighborhoods. Suppose the resulting differentiable objective is lower bounded by $\mathcal L_{\inf}$ and has block Lipschitz constants $L_\theta,L_\psi>0$. Exact block-gradient steps of size $1/L_\theta$ and $1/L_\psi$ satisfy
\begin{equation}
\mathcal L^t-\mathcal L^{t+1}\ge
\frac{\|g_\theta^t\|_2^2}{2L_\theta}+
\frac{\|g_\psi^t\|_2^2}{2L_\psi}.
\end{equation}
Summing over $T$ rounds bounds the average block-update squared gradient norms by $2L_{\max}(\mathcal L^0-\mathcal L_{\inf})/T$. The result applies to the smoothed objective, or locally away from the kinks of the positive-part map in Eq.~\ref{eq:routing}.

\section{Evaluation Protocol}
\label{app:metrics}

\textbf{Attribute localization.}
Let $F=\{i:y_i=1\}$ be held-out failures, $P=\{i:p_i\ge t\}$ predicted failures, and $G_i\subseteq\C_v$ the independently validated repair set. Let $T_i^k$ denote the top-$k$ non-null predicted attributes. Attribute hit rate is
\begin{equation}
\mathrm{Hit@}k=\frac{1}{|F|}\sum_{i\in F}\ind[T_i^k\cap G_i\ne\varnothing].
\end{equation}
Candidate omissions and abstentions remain in the denominator. For the joint metric, let
\begin{equation}
C=\sum_{i\in F\cap P}\ind[T_i^1\cap G_i\ne\varnothing],\qquad
\mathrm{JF1}=\frac{2C}{|P|+|F|}.
\end{equation}
A detected failure attributed to the wrong asset therefore contributes both a false positive and a false negative under the joint decision.

\textbf{Failure detection.}
Failure is the positive class. AUPRC is computed as non-interpolated average precision on the fixed outcome-labeled mixture, and AUROC is reported where both classes are present. Missing signal sources remain input masks rather than removing tasks. Calibration and threshold selection use only the development partition and are frozen before test scoring.

\textbf{Independent repair labels.}
Repair locations are validated independently from method predictions. Each registered replacement is compared with the unchanged configuration on three matched execution seeds using the official evaluator. An attribute is accepted when the matched executions provide consistent evidence of failure-to-success repair without an opposing success-to-failure transition. Borderline cases are resolved through a prediction-blind adjudication workflow that inspects the matched baseline/replacement trajectories, official action checks, environment assertions, and the intended replacement mechanism. These labels are used only for evaluation and are excluded from model fitting, threshold selection, and intervention ranking.

\textbf{Executed repair metrics.}
Let $F_{\mathrm{nat}}$ and $S_{\mathrm{nat}}$ denote the natural failure and success cohorts. For execution repeat $r$, Failure Recovery is the fraction of tasks in $F_{\mathrm{nat}}$ that succeed after the selected repair, Clean Regression is the fraction of tasks in $S_{\mathrm{nat}}$ that fail after applying the same policy, and Overall Success is the success rate over all 114 evaluation tasks. Principal repair results use three matched repeats shared across methods. We first compute one cohort-level metric per repeat and then report the mean and sample standard deviation across the three repeats.

\textbf{Matched execution seeds.}
The three execution seeds are deterministically derived per $(\text{base task},\text{repeat})$ rather than using three global seeds. A fixed final seed $2026092601$ is hashed into a base seed, and each task/repeat pair is mapped through the same deterministic paired-seed rule. All compared methods therefore encounter identical task--repeat seed pairs.

\textbf{Paired confidence intervals.}
We use 10,000 paired base-task bootstrap resamples with random seed 20260926 and percentile intervals. For each task, the three matched outcomes are first averaged; the resulting case-level treatment--reference difference is then resampled by base task, keeping all matched outcomes of a sampled task together. Reported 95\% intervals are the 2.5th and 97.5th percentiles of the paired bootstrap distribution.

\textbf{Training-time conditional effects.}
For a candidate second edit $c$ and conditioning set $S_0\subseteq\C(z)$, the training-side conditional effect averages the marginal contribution of $c$ over replay steps in which every attribute in $S_0$ already appears in the current prefix:
\begin{equation}
\Delta(c\mid S_0)
=
\frac{1}{N_c}
\sum_{q,\pi,r}
\sum_{k:\,S_0\subseteq P_{\pi,k}}
\left[
r_q(P_{\pi,k}\cup\{c\})-r_q(P_{\pi,k})
\right],
\label{eq:conditional-effect}
\end{equation}
where $P_{\pi,k}$ is the set of replacements already applied before step $k$ in permutation $\pi$. Valid marginal samples are pooled across training tasks, permutations, and repeats. Attributes in $S_0$ need not be immediately adjacent to $c$; they only need to precede $c$ in the sampled replacement order. In the interaction-transfer experiment of Section~\ref{sec:interaction}, $S_0=\{\mathrm{Skill}\}$. Candidate second edits are ranked from this training statistic before any held-out repair execution.

\textbf{Offline replay and test-time diagnosis.}
The 5,280 CTCA executions are training-time supervision. At evaluation time, DAAF predicts repair effects from the observed execution and registered assets rather than replaying every candidate replacement. Final repair executions are used only to measure the outcome of the selected policy.

\section{Additional Experimental Evidence}
\label{app:additional-results}

\subsection{Extended Attribute-Localization Controls}
\label{app:extended-localization}

Table~\ref{tab:extra-localization} reports two additional references for the attribute-localization experiment. Uniform candidate choice averages over the accepted repair sets and can therefore exceed $20\%$ when a failure admits more than one validated repair. LLM + CTCA retains the trajectory-based LLM proposal interface while adding CTCA as a partial component rather than using the complete DAAF decision rule.

\begin{table}[htbp]
\centering
\caption{Additional attribute-localization controls on the 83 natural Telecom failures.}
\label{tab:extra-localization}
\small
\begin{tabularx}{0.72\linewidth}{@{}Ycc@{}}
\toprule
Method & Attribute Hit@1 $\uparrow$ & Joint F1 $\uparrow$ \\
\midrule
Uniform candidate choice & 37.35 & -- \\
Component-aware LLM & 40.96 & 51.13 \\
LLM + CTCA & 34.94 & 46.03 \\
Full \method{} & \textbf{80.72} & \textbf{68.02} \\
\bottomrule
\end{tabularx}
\end{table}

The partial LLM + CTCA composition does not reproduce the gain of the full decision pipeline. This result is consistent with the role of DAAF as a joint detection-and-attribution procedure rather than a replacement-effect scorer attached to an unchanged proposal rule.

\subsection{Detection-Only Control}
\label{app:detection-only}

The detection-only control uses the same registry, candidate budget, routing rule, neighborhood construction, training length, learning rate, temperature, and evaluation protocol as Full DAAF. The detector is trained normally, while the typed attribution parameters remain fixed at initialization and receive no replacement-effect updates. The control therefore tests whether failure recognition alone can support the downstream repair decision.

\begin{table}[htbp]
\centering
\caption{Detection-only control under the three-repeat Telecom repair protocol. $\Delta$ Recovery is measured relative to Full DAAF.}
\label{tab:detection-only}
\small
\begin{tabularx}{0.82\linewidth}{@{}Yccc@{}}
\toprule
Method
& Failure Recovery $\uparrow$
& $\Delta$ Recovery 95\% CI
& Clean Regression $\downarrow$ \\
\midrule
Detection-only control
& $4.42\pm2.51$
& $-58.23\ [-66.27,-49.80]$
& $24.73\pm1.86$ \\
Full \method{}
& $\mathbf{62.65\pm3.19}$
& reference
& $\mathbf{3.23\pm3.23}$ \\
\bottomrule
\end{tabularx}
\end{table}

The paired interval excludes zero by a wide margin, showing that failure detection without learned replacement effects is insufficient for the repair policy used by DAAF.

\subsection{Interaction Statistics in the Replay Corpus}
\label{app:interaction-stats}

Table~\ref{tab:interaction-stats} reports the per-attribute activity statistics underlying Section~\ref{sec:interaction}. ``Standalone'' applies an attribute to the original configuration, whereas ``Composed'' measures the same attribute after other registered changes have modified the system state.

\begin{table}[htbp]
\centering
\caption{Replacement activity on 4,400 controlled replay steps from 220 CTCA training tasks. Values are percentages of replacement steps that change official task reward.}
\label{tab:interaction-stats}
\small
\begin{tabularx}{0.82\linewidth}{@{}Yccc@{}}
\toprule
Attribute & Standalone & Composed & Change \\
\midrule
Harness: parallel tool calls & 0.62 & 30.85 & +30.23 \\
Knowledge: main policy & 0.53 & 30.00 & +29.47 \\
Prompt: system template & 1.25 & 27.42 & +26.17 \\
Prompt: instruction & 0.53 & 21.34 & +20.81 \\
Skill: support procedure & 52.20 & 45.35 & -6.85 \\
\midrule
All attributes & 11.36 & 31.02 & $+19.66\ [14.89,24.43]$ \\
\bottomrule
\end{tabularx}
\end{table}

Across the replay corpus, 515 task--attribute pairs exhibit a measurable effect somewhere in their replacement chain, while standalone probing identifies only 59. Thus isolated probing misses 88.54\% of the observed effective pairs. For 472 of the 1{,}100 task--attribute targets, sampled marginal contributions differ in sign across replacement orders. These statistics support modeling signed marginal effects in composed intervention contexts rather than treating each edit as having a fixed isolated effect.

\subsection{Incomplete-Evidence Analysis}
\label{app:incomplete-evidence}

Table~\ref{tab:missingness-app} compares three treatments for unavailable failure signals. Missingness-aware encoding keeps explicit availability indicators; mean imputation uses the training-side observed mean for an unavailable source; zero fill substitutes an observed value of zero.

\begin{table}[htbp]
\centering
\caption{Failure detection under incomplete evidence (\%, higher is better).}
\label{tab:missingness-app}
\small
\begin{tabularx}{0.82\linewidth}{@{}Ycccc@{}}
\toprule
& \multicolumn{2}{c}{Natural cohort}
& \multicolumn{2}{c}{Balanced development} \\
\cmidrule(lr){2-3}\cmidrule(l){4-5}
Treatment & AUPRC & AUROC & AUPRC & AUROC \\
\midrule
Missingness-aware (\method{}) & \textbf{75.77} & \textbf{56.06} & \textbf{66.00} & \textbf{73.92} \\
Mean imputation & \textbf{75.77} & \textbf{56.06} & \textbf{66.00} & \textbf{73.92} \\
Zero fill & 71.31 & 44.58 & 50.00 & 48.00 \\
\bottomrule
\end{tabularx}
\end{table}

Missingness-aware encoding and mean imputation are identical on the two reported splits, so these experiments do not establish an advantage of explicit masks over mean imputation. The supported conclusion is narrower: treating an unavailable source as an observed zero degrades failure detection, especially on the balanced development set. DAAF keeps availability explicit to avoid conflating missing evidence with observed low-failure evidence.

\subsection{Neighborhood Quality under Attribute Hold-Out}
\label{app:neighborhood-quality}

For each of the $\binom{5}{3}=10$ attribute splits, three attributes are used to construct the graph and the remaining two are held out for evaluation. For each selected edge $(i,j)$, held-out agreement is one when the two endpoints have the same top-response attribute among the two held-out attributes:
\[
\mathrm{Agree}(i,j)
=
\ind\!\left[
\arg\max_{c\in H}\widehat\phi^{\mathrm{MC}}_{i,c}
=
\arg\max_{c\in H}\widehat\phi^{\mathrm{MC}}_{j,c}
\right],
\]
where $H$ is the held-out attribute pair. Intervention-response neighborhoods use the kernel in Eq.~\ref{eq:neighbors-app} with $\ell=0.2$ and retain five neighbors per node. Request-only neighborhoods use cosine similarity of the fixed 64-dimensional request representation, and Random uses seed-fixed random edge selection.

\begin{table}[htbp]
\centering
\caption{Neighborhood quality under attribute hold-out, averaged across all ten splits.}
\label{tab:neighborhood-app}
\small
\begin{tabularx}{0.72\linewidth}{@{}Ycc@{}}
\toprule
Construction & Held-out Agreement $\uparrow$ & Splits Won \\
\midrule
Intervention response (\method{}) & \textbf{62.63} & \textbf{10 / 10} \\
Request similarity only & 60.09 & 0 / 10 \\
Random edges & 58.59 & 0 / 10 \\
\bottomrule
\end{tabularx}
\end{table}

The intervention-response graph has the highest held-out agreement on every split. The attribute hold-out design tests whether the graph captures transferable repair-response structure beyond the attributes used to construct the neighborhood, rather than only lexical request similarity.

\subsection{Online Execution Cost}
\label{app:online-cost}

Table~\ref{tab:online-cost} reports diagnosis- and repair-time environment executions. The offline CTCA replay corpus is excluded because it is shared training supervision rather than a per-request diagnosis cost.

\begin{table}[htbp]
\centering
\caption{Online environment executions per natural failure. Lower is better.}
\label{tab:online-cost}
\small
\begin{tabularx}{0.62\linewidth}{@{}Yc@{}}
\toprule
Method & Online Runs / Failure $\downarrow$ \\
\midrule
No-edit Rerun & 1.00 \\
Component-aware LLM & 1.48 \\
LLM + MSWSD & 2.00 \\
Full \method{} & 2.00 \\
\bottomrule
\end{tabularx}
\end{table}

Full DAAF does not enumerate the five candidate replacements online. Its controlled intervention budget is paid once during training; the online count reflects only the diagnosis and selected-repair protocol used in evaluation.

\section{Experimental Settings}
\label{app:settings}

\subsection{Task Partitions and Supervision}
\label{app:task-partitions}

The Telecom evaluation set contains all 114 official default tasks grouped by canonical base-task identity, yielding 83 natural failures and 31 natural successes under the frozen evaluation configuration. A disjoint 50-task development set contains 25 failures and 25 successes and is used only for model selection and calibration. Training partitions exclude all evaluation and development base-task identities together with their repeated executions and intervention variants.

\textbf{MSWSD training pool.}
The final MSWSD training pool contains 476 evaluator-anchored trajectories from 395 base tasks, with 327 failure-oriented and 149 success-oriented seeds. Available sources include trajectory anomaly, visible user feedback, and repeated-run disagreement when matched observations exist. All signal extraction rules are frozen before evaluation.

\textbf{CTCA replay corpus.}
CTCA uses 220 controlled tasks with five registered candidates: two Prompt attributes and one each from Harness, Knowledge, and Skill. Two replacement permutations and two matched repetitions require 24 valid replay executions per task and provide 1,100 signed marginal-effect targets in total. The same replay corpus is used to estimate the conditional effects in Eq.~\ref{eq:conditional-effect}; held-out outcomes are not consulted when ranking the evaluated compositions.

\textbf{Registry correspondence for transfer experiments.}
All interventions remain versioned through the registry. When a training-time intervention role is evaluated under the held-out system configuration, correspondence is established through the registered component type, canonical source role, interface and dependency constraints, and replacement-policy identity before test execution. The transfer experiment applies the evaluation registry's registered replacement for the selected role; only the training-side conditional-effect statistic determines the ordering used in the held-out comparison.

\textbf{Models.}
Telecom execution, user simulation, and the component-aware LLM diagnostician use GPT-5.6 Luna~\citep{openai2026gpt56}. Who\&When localization uses DeepSeek-V4-Flash. Model and execution identities are retained with the experiment artifacts.

\subsection{Optimization and Model Selection}
\label{app:optimization-settings}

MSWSD uses full-batch optimization with learning rate $0.2$ and three base-task folds for cross-fitted weak supervision. CTCA and the full joint model use full-batch gradient descent with learning rate $0.2$ and global gradient-norm clipping at $5.0$. The final configuration is reported in Table~\ref{tab:full-config}. Thresholds and calibration parameters are selected on the 50-task development configuration and frozen before evaluation.

\textbf{Detection-only control.}
The detection-only control uses the same training seed, candidate budget, neighborhood rule, temperature, routing epsilon, number of epochs, warm-up, learning rate, and evaluation protocol as Full DAAF. Its typed attribution parameters remain frozen at initialization and receive no replacement-effect update. The detector threshold is selected on the same development configuration and is $0.7440$.

\subsection{Matched-Seed Repair Evaluation}
\label{app:matched-seeds}

The principal Telecom repair comparisons use three matched execution repeats for every evaluation task. The final seed identifier is 2026092601. A deterministic hash produces a common base seed and then derives one seed for each $(\text{base task},k)$ pair, $k\in\{0,1,2\}$. The same task/repeat seeds are reused across compared methods, enabling paired outcome analysis while keeping different tasks statistically distinct.

\subsection{Who\&When Reporting}
\label{app:whowhen}

Who\&When contains 126 Algorithm traces and 58 Handcrafted traces. Agent, Step, and Joint accuracy use the benchmark's native labels with common explicit-author-name normalization and unchanged step indices. The DAAF frontend is evaluated with the same trace inputs as the reported baselines. Because Who\&When does not expose editable system assets and the compared systems are not presented as a model-matched repair evaluation, we use it only as an auxiliary localization check. The executable repair claims in the main text are established by the Telecom study.

\end{document}